\documentclass[letterpaper, 10 pt, conference]{IEEEtran} 
\IEEEoverridecommandlockouts                             
\usepackage{hyperref}
\hypersetup{
    colorlinks=true,
    linkcolor=blue,
    filecolor=blue,      
    urlcolor=blue,
    citecolor=blue,
    pdfpagemode=FullScreen,
    }
\usepackage[utf8]{inputenc}
\usepackage{tikz}
\usepackage{cite}
\usepackage[sort,numbers]{natbib}
\usepackage{graphicx}
\usepackage{algorithm}
\usepackage{algpseudocode}
\usepackage[letterpaper,margin=.75in]{geometry}

\title{\LARGE \bf
\vspace{.25in}
Achieving mouse-level strategic evasion performance using real-time computational planning
}

\author{German Espinosa$^{1}$, Gabrielle E. Wink$^{2}$, Alexander T. Lai$^{3}$, Daniel A. Dombeck$^{4}$ and Malcolm A. MacIver$^{4}$
\thanks{*This work was supported by the National Science Foundation, ECCS-1835389 and IIS-2123725 to Malcolm A. MacIver and Daniel A. Dombeck}
\thanks{$^{1}$German Espinosa, Department of Computer Science, Northwestern University, USA;
        {\tt\small germanespinosa@u.northwestern.edu}}%
\thanks{$^{2}$Gabrielle E. Wink, Department of Mechanical Engineering, Northwestern University, USA;
        {\tt\small g-wink@u.northwestern.edu}}%
\thanks{$^{3}$Alexander T. Lai, Department of Biomedical Engineering, Northwestern University, USA;
        {\tt\small alexlai@u.northwestern.edu}}%
\thanks{$^{4}$Daniel A. Dombeck, Department of Neurobiology, Northwestern University, USA;
        {\tt\small d-dombeck@northwestern.edu}}%
\thanks{$^{5}$Malcolm A. MacIver, Center
for Robotics and Biosystems,  Northwestern University, USA;
        {\tt\small maciver@northwestern.edu}}%
}

\begin{document}
\maketitle
\thispagestyle{empty}
\pagestyle{empty}
\begin{abstract}



Planning is an extraordinary ability in which the brain imagines and then enacts evaluated possible futures. Using traditional planning models, computer scientists have attempted to replicate this capacity with some level of success but ultimately face a reoccurring limitation: as the plan grows in steps, the number of different possible futures makes it intractable to determine the right sequence of actions to reach a goal state.  
Based on prior theoretical work on how the ecology of an animal governs the value of spatial planning, 
we developed a more efficient biologically-inspired planning algorithm, \emph{TLPPO}. This algorithm allows us to achieve mouse-level predator evasion performance with orders of magnitude less computation than a widespread algorithm for planning in the situations of partial observability that typify predator-prey interactions. We compared the performance of a real-time agent using \emph{TLPPO} against the performance of live mice, all tasked with evading a robot predator.
We anticipate these results will be helpful to planning
algorithm users and developers, as well as to areas of neuroscience where robot-animal interaction can provide
a useful approach to studying the basis of complex behaviors.


\end{abstract}

\section{INTRODUCTION}

Animals equipped with planning abilities present rich and complex behaviors. From big cats stalking antelopes on the savanna to mice using clutter to escape foxes in a forest, animals show the extraordinary advantages of this action-control mechanism when dealing with life-and-death situations, where trial and error learning easily transitions to trial and death.  

Prior theory and simulations work in paleontology and neuroscience have provided evidence that planning may confer a ``selective benefit''---necessary for the adoption of a trait through a population over evolutionary time---in predator-prey interactions in patchy savanna-like terrestrial domains \cite{MacI09a,MacI17a,Muga20a,MacI22a}.
During these encounters, animals need to keep track of the continuously changing state of the world affected by every move of both animals. They also must estimate information beyond what their senses can provide, as the adversary could be temporarily out of sight. Furthermore, they create predictions about the future to determine the outcome of possible actions. This process is done in real-time with extraordinary efficiency. All of these characteristics make animal planning in naturalistic scenarios significantly more complex than many of the tasks 
considered in contemporary applications of online planning within AI \cite{Wang20a,Mazz20a,Farh21a, Xia19a, Gold14a, Fede20a, Coue11a, Sunb18a, Mern20a, Gal13a}. 
\begin{figure}[t!]
	\centering
	\includegraphics[width=.95\columnwidth,
	keepaspectratio]{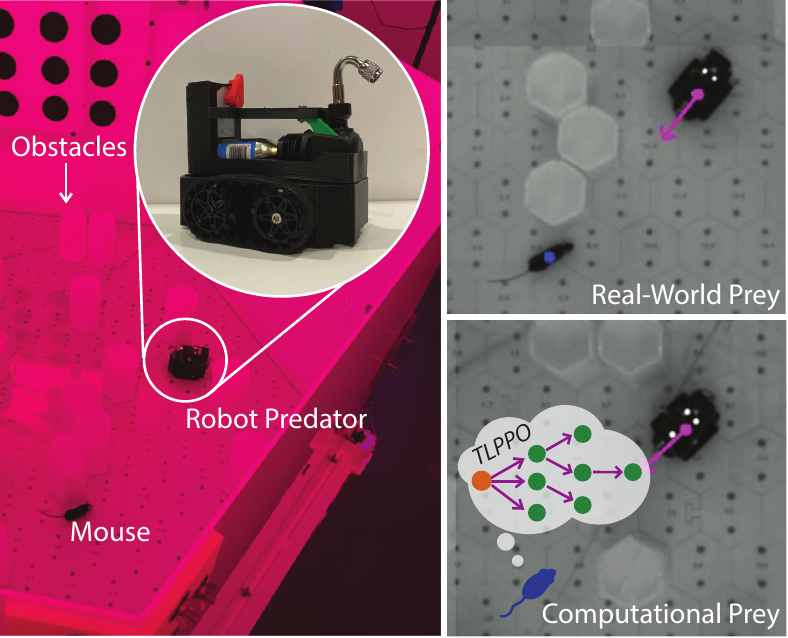}
	\caption{\footnotesize Predator-prey interaction in a configurable environment with a robotic predator (\textbf{left}) pursuing a live mouse (\textbf{right top}) or a computational agent (\textbf{right bottom}) -- see video \href{https://youtu.be/PpGYyq2HbUY}{https://youtu.be/PpGYyq2HbUY}. 
}\normalsize
	\label{experimental_setup}
\vspace{-0mm}
\end{figure}

The present work aims to reduce the task complexity gap between AI and biological planning by developing a computational agent capable of strategic evasion in real-time at the same level as live animals. For this purpose, we used a novel, ethologically relevant task: a predator-prey interaction where lab mice play the role of the prey, and a reactive robotic agent plays the role of a predator. 
In this task, a mouse traverses a large semi-occluded arena to reach a reward while avoiding an aversive air puff given by a roaming robot predator. The mouse's performance was quantified by the percentage of runs where the mouse completed the task without receiving the predator's puff. This proxy for survival rate served as the target for a computational agent performing the same task. 

The state-of-the-art planning algorithm Partially Observable Monte-Carlo Planning (\emph{POMCP}) was initially used for the development of the agent \cite{Silv10a}. \emph{POMCP} is a comprehensive framework for planning in partially observable scenarios. It provides both a mechanism to create and maintain a statistical representation of the current state beyond the observable world (belief state) and a configurable high-performance tree-search algorithm that balances the exploration-exploitation trade-off to determine the best next action. 
To test the performance of the agent, a simulation replicating the experiment was created. 
However, during this test, it became apparent that the computational effort required to reach the target performance exceeded, by several orders of magnitude, the time available available to select moves in the real experiment. This inefficiency made the \emph{POMCP} agent not performant enough for real-time execution and made it evident that a new approach was needed.

To address this problem, we propose a new spatial planning algorithm inspired by prior evolutionary computational neuroscience results \cite{Muga20a}, ``Trajectories over Locations where Planning Pays Off'' or \emph{TLPPO}. \emph{TLPPO} uses the properties of the graph structure of the space where the predator-prey interactions occur to determine places that might contain opportunities. Then, during the planning phase of the agent, it forces the tree-search process to explore branches in the direction of those opportunities. Simulation results using \emph{TLPPO} showed a significant decrease in the computational effort required to compute plans that match animal survival rates. To test the \emph{TLPPO} agent in the same conditions that were used to determine animal performance, we repeated the mice experiment with the exact same configuration (same average speed as the mice shown in trials, same robotic predator), but this time we replaced the live animals with the \emph{TLPPO} agent. 

Our results show that controlling the actions of our planning agent through the \emph{TLPPO} algorithm enabled the agent to interact with an aggressive robotic predator in real-time successfully. Its performance was comparable to a live mammal, for which there is growing neurophysiological and behavioral evidence of planning.

\section{Previous Work}

In the last decade, online planning research has mainly been addressed by online partially observable
Markov Decision problem (POMDP) solvers \cite{Wang20a,Mazz20a,Farh21a, Xia19a, Gold14a, Fede20a, Coue11a, Sunb18a, Mern20a} based on the Partially Observable Monte Carlo Planning (\emph{POMCP}) \cite{Silv10a} algorithm. \emph{POMCP} was first introduced in 2010 and is still considered the fastest online planning POMDP solver to date. However, its implementation comes with some caveats. 

The most important challenge of using \emph{POMCP} is the exponential growth of the decision tree while performing the forward information search. This problem is known as the ``curse of dimensionality.'' Even with relatively small problems, the decision tree cannot be explored exhaustively. \emph{POMCP} addresses this problem by using the Upper Confidence Bound 
(\emph{UCB1})
algorithm \cite{Brow12b}. During the tree-search phase, \emph{UCB1} stores statistics (visit count and reward values) for each branch expanded in the past and uses them to determine what branches to expand in the future. This algorithm can be configured to 
balance ``exploration''---expanding branches that had been less visited before---and ``exploitation''---expanding branches that had better reward values in the past. 
This makes \emph{UCB1} effective when rewards can be obtained quickly, and the tree has a gradient of reward values that grows towards the optimal plan. But in cases with sparse rewards that require a large number of steps to obtain them, the computational effort required to effectively plan is extremely large. 

A commonly used approach to overcome the curse of dimensionality is artificially limiting the number of actions available for tree expansion and increasing it progressively, called Double Progressive Widening (\emph{DPW}) \cite{Coue11a,Sunb18a,Mern20a}. By doing this, a smaller section of the tree is tested more frequently, and as a result, planning performs a smaller number of policy reward estimations but with higher certainty due to the increased repetitions. \emph{DPW} can be used in trees with massive expansion (even in continuous spaces), but it still relies on being able to find a gradient of reward values in every direction. With sparse rewards, \emph{DPW} has a small probability of reaching them, and policy estimation suffers. A different approach to tackle the curse of dimensionality is to memorize and reuse parts of the plan computed previously, called Adaptive Belief Tree (\emph{ABT}) \cite{Kurn16a}. This algorithm stores the policies computed in previous planning iterations and determines when the environment has changed enough to motivate a recalculation. \emph{ABT} is extremely useful when the planning process occurs faster than the changes in the environment or when the environment is highly predictable. However, in unpredictable dynamic environments, like dealing with a mobile adversary, policies become obsolete faster than they can be computed. 

\section{The \emph{TLPPO} Algorithm}
As discussed above, current sub-sampling methods depend mostly on the reward distribution to provide clues about where the branches of the decision tree with higher values are. But in a highly volatile environment where the rewards are sparse, and the required number of steps to obtain them is large, these approaches fail to build a correct policy value estimation in a reasonable time. 
Here we propose a method inspired by neuroscience research to use additional information that is available to the agent to inform the tree expansion.

\begin{figure*}[t!]
	\centering
	\includegraphics[width=.9\textwidth,
	keepaspectratio]{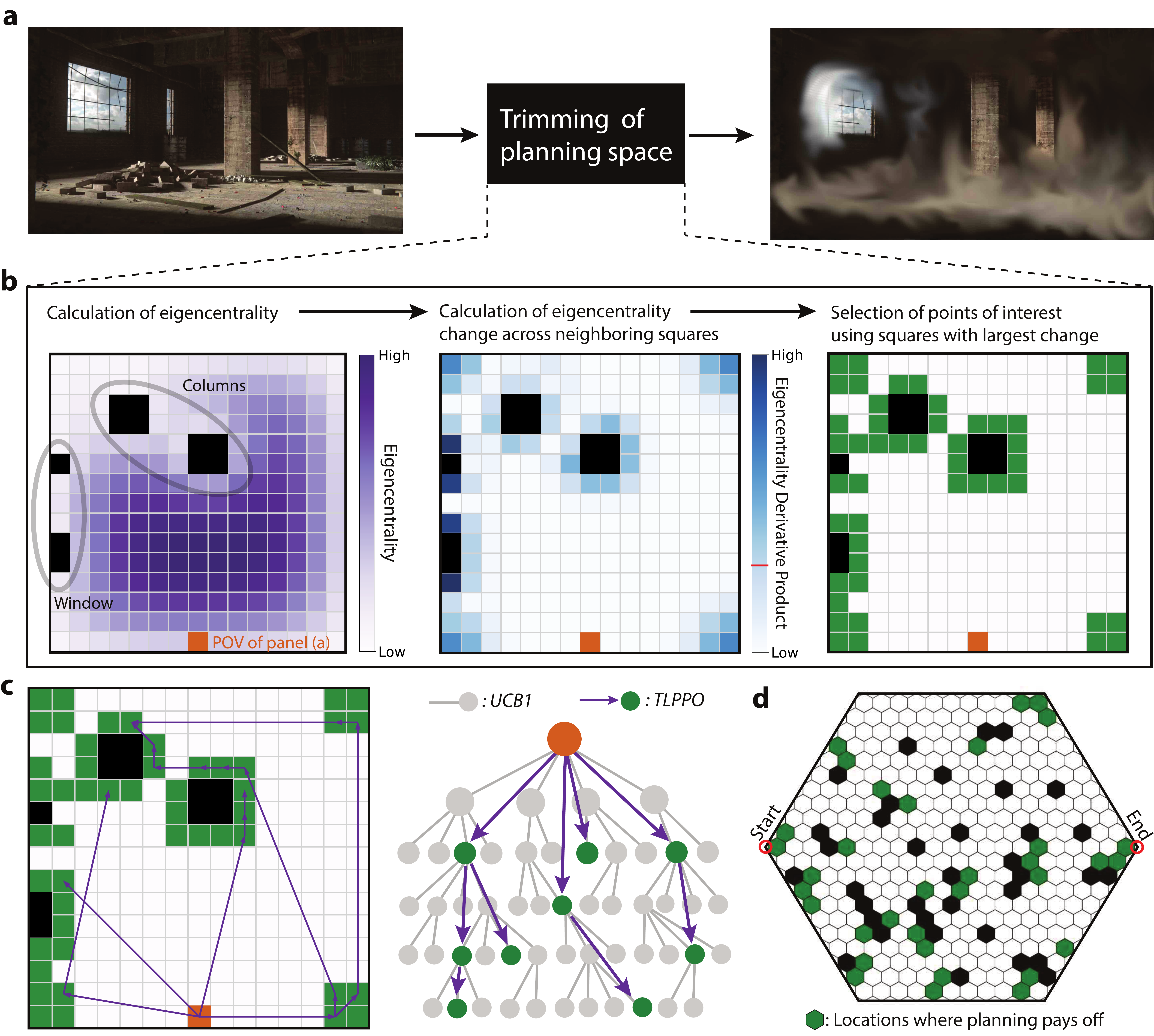}
	\vspace{-0mm}
	\caption{\footnotesize
\emph{TLPPO}'s selection of locations where planning pays off greatly reduces  
the number of locations that need consideration in a decision tree, enabling
real-time performance.  (\textbf{a}) 
	Trimming of features in the space.
Only useful features during an adversarial interaction (window,
columns) are highlighted, as identified
by the value of their eigencentrality
in the 2D projections shown below.
(\textbf{b})
Workflow of \emph{TLPPO} algorithm. In a 2D
map of the example space, eigencentrality of each square (purple, left) is computed. 
Derivatives along each direction is computed, and
the product of these derivatives
(blue, center) is calculated. Squares that exceed a threshold (red line in the colorbar) are selected as locations where 
planning pays off (green, right).
(\textbf{c})
Example of tree exploration using \emph{TLPPO}.
The agent only evaluates trajectories connecting locations where planning
pays off
using the shortest path between them (left). Instead of expanding
new trajectories after every action as occurs in Monte-Carlo tree
search, the algorithm
only expands when it reaches a location
where planning pays off, and only in the
direction of other such locations visible from its current
location (right). (\textbf{d}) Green
hexagons show locations where planning pays off in a schematic of the same 
hexagonal arena as the predator-prey experiments. Black cells are occlusions blocking movement and vision.
}\normalsize
	\label{poi_prelim}
	\vspace{0mm}
\end{figure*}

\vspace{3mm}
\noindent\textbf{Connectedness and the cognitive map.}
One of the key assumptions used in testing different hypotheses about biological spatial planning is that the brain has a representation of the space around it. 
There is evidence that this representation is formed in a brain structure called the hippocampus, where neurons called \textit{place cells} get associated with specific locations in space, thereby forming a cognitive map \cite{Keef78a}. 
Using the cognitive map, the brain is thought to identify connections between locations and possibly provide the basis for simulating different trajectories in imagination. These trajectories have been recorded in animal brains during planning behaviors using neural probes \cite{Schm18a, Kay20a}. 

The mice used in our experiments were allowed ``practice rounds'' to explore the arena without the presence of the predator robot to let them memorize the space and build their cognitive maps. The computational agent receives the coordinates of the occlusions at the beginning of each experiment to build its own version of it. Cognitive maps are powerful tools by themselves; however, evidence suggests that animals might use additional information to enrich them.
Previous research \cite{Muga20a} proposed the animal brain might have evolved mechanisms to identify aspects of the graph structure of its habitat to determine connectedness of the space. Animals can use this information to determine the difference between open and closed spaces. This could provide clues about how safe the current state of the world is (e.g., prey in an open space can be seen from far away by a predator). Computationally, connectedness of the space can be estimated using a graph theory metric called eigenvector centrality (hereafter, eigencentrality) \cite{Muga20a}. Eigencentrality shows not only how a portion of the space is connected to its neighbours but also how well connected its neighbors are to their neighbors and so on.

In prior work on arbitration between model-free and model-based action control, we showed that eigencentrality can be used to activate planning when moving from a closed to an open space without decreasing survival rate relative to much more expensive full-time planning \cite{Muga20a}. This suggests that the geometry of the space influences the value of planning. Specifically, planning only generates a benefit when performed in places with a high gradient of connectedness (transitional spaces from closed to open). We call these places ``locations where planning pays off.'' To illustrate how these locations are distributed in the environment, we will use a scene from an abandoned factory (Fig. \ref{poi_prelim}\textbf{a} left). The factory contains a broken window, piles of bricks, timber pieces, rocks, columns and considerable open space between them. However, if a human entered this factory during a life-and-death situation (e.g. an active shooter) and had to plan an escape, there are only two useful features: the broken window and the columns (Fig. \ref{poi_prelim}\textbf{a} right). These features 
afford two critical behaviors: 
fleeing and hiding. In the 2D representation of the connectedness of the same space (Fig. \ref{poi_prelim}\textbf{b} left), those same areas present the highest fluctuations of eigencentrality. To isolate the corresponding cells, we compute these fluctuations as derivatives (one for each possible direction) and store their product as an intermediate step. This ensures the eigencentrality change is occurring in every direction (Fig. \ref{poi_prelim}\textbf{b} center). Finally, it is only necessary to apply a threshold; cells with stored values above or equal to the threshold are defined as locations where planning pays off (Fig. \ref{poi_prelim}\textbf{b} right).
Figure \ref{poi_prelim}\textbf{d} shows the result of this process with the habitat configuration used for the experimental work.

\vspace{3mm}
\noindent\textbf{Trajectories over locations where planning pays off.} After identifying the locations where planning is beneficial, the agent still has to manage the full computational burden of performing the planning process. To address this problem, we use the identified locations during the tree-search phase to create a new algorithm ``Trajectories over Locations where Planning Pays Off'' (\emph{TLPPO}). \emph{TLPPO}  assumes that in locations outside of those with high potential pay off, according to the definition above, there will also be no benefit from expanding the tree while planning. This means that tree expansions should only happen at locations where planning pays off and only in the direction of other locations where planning pays off using the shortest path between them (Fig. \ref{poi_prelim}\textbf{c}). This creates a subset of trajectories to explore that is several orders of magnitude fewer than the total universe of possible trajectories. Reducing the number of action sequences to simulate during planning in this manner has the potential benefit of accelerating and simplifying the process greatly.

\section{Mammalian-level performance baseline}

Assessment of an animal's strategic evasion behavior and performance was done in a new naturalistic, predator-prey task whose design was informed
by prior theory, paleontological and evolutionary neuroscience findings, and simulations 
\cite{MacI09a,MacI17a,Muga20a,MacI22a}.
The task consists of a regular hexagonal arena with a long diagonal length of 2.34~m. The arena is overlaid with a hexagonal grid system consisting of 330 magnetic hexagon cells
(Fig.~\ref{experimental_setup} \& \ref{poi_prelim}\textbf{d}). The distance between two cell centers is 11 cm. Obstacles with magnetic bases are placed in a configuration that maximizes the utility of strategic behavior
from prior results \cite{Muga20a}. A custom robot predator equipped with an air puff mechanism (Fig.~\ref{experimental_setup}), serving as an aversive stimulus in lieu of actual prey 
capture, actively roams the arena in search of the animal subject. The animal's task for each episode is to traverse the arena to a goal location without being captured by the predator. An episode ends if the animal reaches the goal or is captured. A capture is recorded if the distance between the robot and the subject falls below 2.5 cells (27.5~cm), the 
threshold to trigger an air puff. 

\begin{algorithm}
\caption{Autonomous robot predator behavior}\label{alg:cap}
\begin{algorithmic}[1]
\While{experiment is running}
    \State Find spawn cell \emph{S}
    \State Move robot to \emph{S}
    \While{episode is running}
        \If{mouse is visible}
            \State Move robot to last seen mouse cell
        \ElsIf{mouse is not visible}
            \State Find cells not visible to robot 
            \State Randomly select a non-visible cell \emph{N} 
            \State Move robot to \emph{N}
        \EndIf
    \EndWhile
\EndWhile
\end{algorithmic}
\label{predator_algorithm}
\end{algorithm}

\begin{figure*}[t!]
	\centering
	\includegraphics[width=.9\textwidth,
	keepaspectratio]{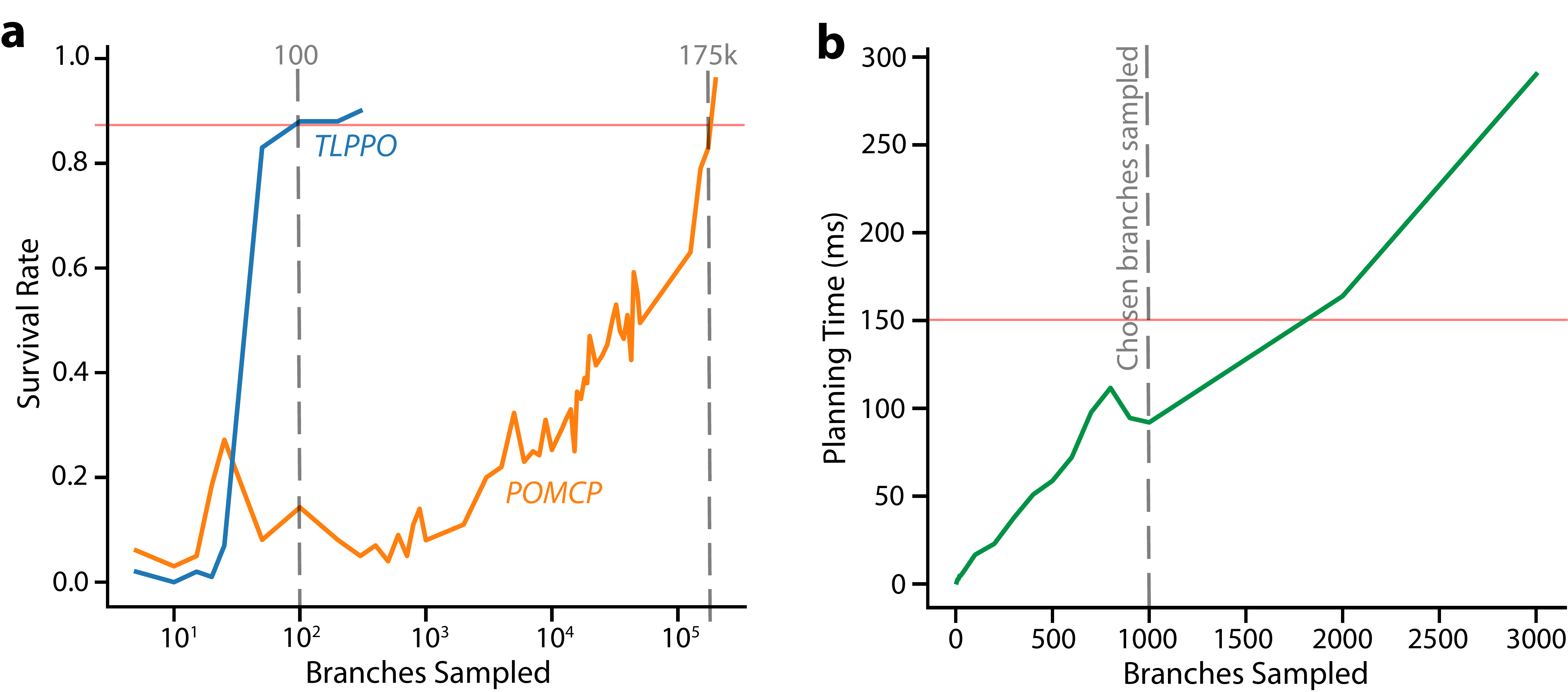}
	\caption{\footnotesize
(\textbf{a}) Survival rate as a function of branches sampled for \emph{TLPPO} and \emph{POMCP} planning algorithms. Red line represents the survival rate calculated from mice experiments. (\textbf{b}) Branches sampled versus time needed. Planning time was the average time across both \emph{POMCP} and \emph{TLPPO} algorithms, which were roughly equal in time per branch sampled. Red line represents the maximum allowable planning time given the speed of the mouse and arena cell size.
}\normalsize
	\label{TLPPO_POMCP}
	\vspace{0mm}
\end{figure*}

The predator starts each episode at a spawn cell: a cell not visible to the prey, in a region of the arena (1/3 of the total area) furthest from the start gate (Fig.~\ref{poi_prelim}\textbf{d}) through
which the animal enters the arena. 
The robot then follows the behaviors described in Algorithm\textbf{ \ref{predator_algorithm}} until the end of the experiment. 
A high-speed four-camera tracking system actively records the position of the robot and the animal, providing real-time control information for the predator and recording animal behavior at 30 Hz. Cameras and encoders are utilized for PID control of the differential drive autonomous predator robot. The robot follows trajectories created by an A* algorithm \cite{Hart68a} used to find the shortest path to the destination with obstacle avoidance. Occasional control intervention with a joystick was applied if the robot navigated into an obstacle and failed to return to its trajectory. 

Two male lab mice (\emph{Mus musculus}, strain C57BL/6) were water restricted and trained in this setup to traverse the arena while actively avoiding the robot. Progression through the arena is controlled by automated doors and water feeders that provide water rewards and motivation for the mouse to run multiple trajectories for a given experiment. Training consisted of a multi-step process where the mouse performed one 30-minute experiment each day and was gradually exposed to the environment without the robot to learn the space and build a cognitive map. Once the mouse has explored the entirety of the arena and performed more than 1 trajectory/minute, the mouse was given multiple days to perform experiments with the robot to learn the robot's behavior. Once the number
of loops through the arena over an experimental session stabilized, data collection began. Each mouse performed 5 days of robot experiments. During both training and experiments, a white noise generator covered the sound of the robot's movements and the arena was cleaned before each experiment to remove olfactory cues, largely limiting the mouse's sensory cues to vision. Both mice were control
subjects of a parallel experiment. As controls, these mice had received a craniotomy and an injection of an inert viral vector before training and received daily intraperitoneal injections of saline during experiments.
All experiments were approved and conducted in
accordance with the Northwestern University Animal Care and Use Committee.

Average mice and robot speeds during movement were calculated by extracting the position of both agents at every frame and then calculating a subsequent array of velocities where ${\Delta x}$ was the change in position and ${\Delta t}$ was the change in time between each frame. To focus on speed primarily during movement and not during pauses, only velocities above 0.15~m/s were selected for both agents. Then, an average across both agents was taken across all remaining velocities and all experiments. The average speed of the mouse was found to be 0.76~m/s.

Baseline survival rate was calculated by dividing the number of trajectories where the mouse successfully evaded the robot without receiving an air puff by the total number of trajectories across all experiments and mice. The mice ran a total of 230 trajectories and had a survival rate of 0.86.

\section{Matching mammalian performance}

The work explained in the previous section provided insights about two key metrics that will determine if an agent can perform at the same level than the mice in the experiments: survival rate and speed.

\subsection{Survival rate}

To determine the planning effort needed to match the mammalian-level survival rate baseline, we developed a simulation of the experiment where the computational agents are in control of the actions executed by a simulated mouse. In the simulation, the predator is provided with the same reactive behaviors as the robotic predator in the real-world experiment. The key difference between the execution of the simulation and the experiment is that the simulation ``waits'' for the agent to make a decision by stopping all world updates. This allowed us to validate the survival rate of the agents as we increased the number of branches sampled per iteration until the simulated rate was similar to the experimentally measured survival rate. 

\vspace{3mm}
\noindent\textbf{Simulation benchmarks.} Figure \ref{TLPPO_POMCP}\textbf{a} shows the evolution of the performance on the tested agents. To reach mouse-level performance, the \emph{POMCP} agent required sampling 175,000 branches on each planning iteration while the \emph{TLPPO} agent required  $\approx$100 branches. 

\subsection{Speed}

Another critical feature of the planning agent is planning speed due to the interactive nature of the task (evade adversary in real-time). This limits the amount of time that agents can use for planning to the interval between moves. To determine this upper-bound on control delay, we divided the distance between cells in the habitat (0.11~m) by the average mouse speed from the recording sessions (0.76~m/s). This sets the total time that can be used for planning 
to $\approx$150~ms. 

To determine how many branches could be sampled within the time limit, we designed an experiment using the hardware that later will be used to run the agent during the real-world trials. In this experiment we increased the number of planning cycles for both \emph{POMCP} and \emph{TLPPO} until we obtained an average planning time of 75\% of this limit to account for additional delays. This works out to 1000 branches sampled in the real-time version of the agents (Fig.~\ref{TLPPO_POMCP}\textbf{b}).

\subsection*{Comparing mice and computational agents}
\begin{figure}[t!]
	\centering
	\includegraphics[width=.95\columnwidth,
	keepaspectratio]{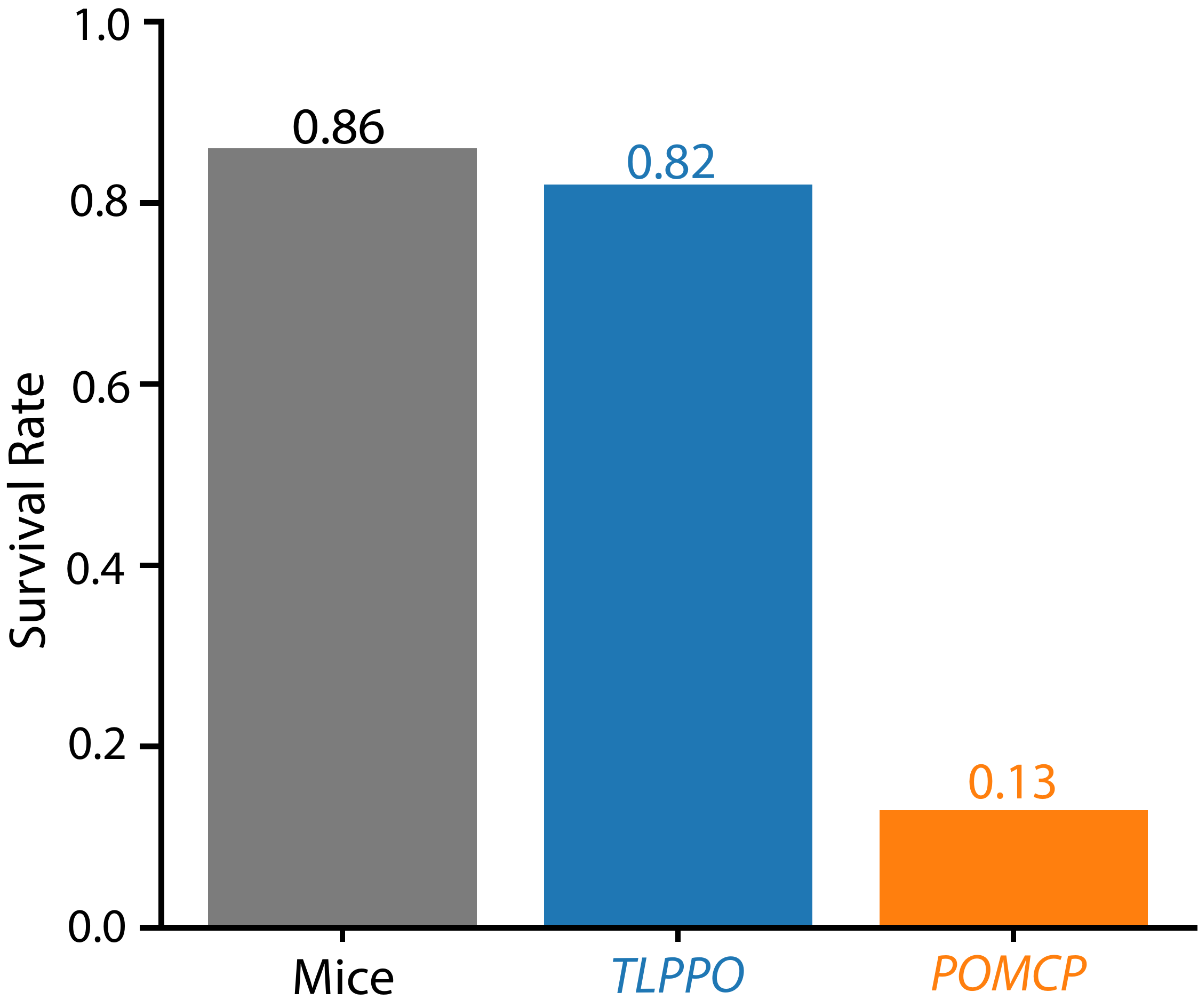}
	\caption{\footnotesize
Survival rate for real-time computational preys performing \emph{TLPPO} and \emph{POMCP} planning algorithms and mice in the presence of a robot predator.  
}\normalsize
	\label{performance_hist}
	\vspace{0mm}
\end{figure}

In the final part of this work, a new set of real-world experiments was executed using the hexagonal arena and the robotic predator. This time the live mice were replaced with the real-time computational agents. For both the \emph{POMCP} and \emph{TLPPO} agents, 100 episodes were recorded to establish their survival rate performance. Results showed that both agents were able to complete task, however, only the \emph{TLPPO} agent was able to obtain a survival rate comparable to the recorded animal performance (Fig.~\ref{performance_hist}).

\section{Discussion} 

We presented a new approach for computational spatial planning in partially observable scenarios that performs on par with live animals. We have shown the efficiency of the \emph{TLPPO} algorithm in comparison to 
\emph{POMCP} and tested its applicability in a real-time complex interactive task with a highly volatile environment.

\vspace{3mm}
\noindent\textbf{Reinforcement learning and model-free agents.}
Here we discuss our use of a model-based method, online planning, rather than other available tools, such as model-free methods (e.g. deep learning approaches). Many current computational approaches
for complex adversarial problems, at least in gaming, rely on model-free reinforcement learning. One important benefit of these techniques is that the agent does not have to perform a tree search at the time it needs to take an action, allowing it to interact with the environment more rapidly. 

One of the major advances in reinforcement learning was the introduction of Deep Q Learning Networks (DQN) \cite{Mnih13a}. The principle behind DQN is that the resulting policy value estimation from a tree exploration done during Q learning can be expressed as a function and, consequently, it can be approximated using deep neural networks. This technique has multiple benefits. One is that DQN allows agents to learn and store information from larger MDP problems. One example is the game of Go \cite{Silv16a,Silv17a}, where the number of possible states ($10^{172}$) is larger than the atoms in the universe, making it impossible to store that information on a traditional Q-table, where every single state requires an entry. Another great benefit is that DQN generalizes the knowledge acquired during training, allowing the agent to make good decisions even in states that have not been visited before. DQN caused a major revolution in AI and became the de facto standard for reinforcement learning with an increasing number of publications using this technique every year \cite{Hest18a,Hess18a,Horg18a,Scha15a,Silv17a,Van2016a,Viny19a,Silv17a,Pan18a,Sali17a}. 

However, all of these solutions require the agent to go through a learning cycle, where it naively explores a problem to learn from experience. This makes
applying them to naturalistic problems challenging, 
where the irreversibility of death looms. In a predator-prey interaction, the prey cannot learn from a failed episode. AlphaGo \cite{Silv16a}, for example, played millions of games before being able to defeat Lee Sedol (world champion of Go at the time) in 2016. This was only possible because it was allowed to gain experience from won and lost games during that training. But this result would have never been possible if the agent had to erase all previous knowledge with every defeat. For these reasons, here we focus on a model-based method only. 
Agents start every episode with nothing gained via experiential learning, but two components of static knowledge: the location of the occlusions and boundaries 
(a cognitive map), and a generative model of the predator behavior as required for evaluating the decision trees. 

There has been a number of prior works exploring interactions between robots and animals \cite{Bie2018a, Rom2018a}. Our study differs from these in that we offer a novel method for directly comparing the behavior, strategy, and performance of a computational planning agent to that of an animal---possibly using similar functions instantiated neurally---performing an identical task.

\section*{ACKNOWLEDGMENTS}
We thank Parker Ryan, Ben Zitzewitz, Lily Browdy, Chris Angeloni, and Sam Griswold for their contributions to the live animal experiments.

\bibliographystyle{ieeetr}
\bibliography{references}

\end{document}